\documentclass[journal]{IEEEtran}
\ifCLASSINFOpdf
\else
\fi
\usepackage{booktabs}
\usepackage{tikz}
\usetikzlibrary{positioning,arrows,fit,backgrounds, patterns}
\usepackage{graphicx}
\usepackage{subcaption}
\usepackage{amsmath}

\begin{document}


\title{Semantic Segmentation of Pathological Lung Tissue with Dilated Fully Convolutional Networks}

% PREVIOUS: Semantic Segmentation of Lung Tissue with Fully Convolutional Networks and Semi-supervised Learning


\author{Marios~Anthimopoulos,~\IEEEmembership{Member,~IEEE,}
        Stergios~Christodoulidis,~\IEEEmembership{Member,~IEEE,}
        Lukas~Ebner,
        Thomas~Geiser,
        Andreas~Christe,
        and~Stavroula~Mougiakakou*,~\IEEEmembership{Member,~IEEE}% <-this % stops a space


\thanks{Manuscript received \today}% <-this % stops a space
\thanks{This research was carried out within the framework of the IntACT research project, supported by Bern University Hospital,``Inselspital'' and the Swiss National Science Foundation (SNSF) under Grant 156511.}% <-this % stops a space
\thanks{M. Anthimopoulos and S. Christodoulidis contributed equally to this work. The asterisk indicates the corresponding author.}% <-this % stops a space
\thanks{M. Anthimopoulos is with the ARTORG Center for Biomedical Engineering
Research, University of Bern, 3008 Bern, Switzerland, and the Department of Emergency
Medicine, Bern University Hospital ``Inselspital'', 3010 Bern, Switzerland
(e-mail: marios.anthimopoulos@artorg.unibe.ch).}% <-this % stops a space
\thanks{S. Christodoulidis is with the ARTORG Center for Biomedical Engineering
	Research, University of Bern, 3008 Bern, Switzerland (e-mail:
	stergios.christodoulidis@artorg.unibe.ch).}% <-this % stops a space
\thanks{T. Geiser is with the University Clinic for Pneumonology, Bern University Hospital ``Inselspital'',
3010 Bern, Switzerland (e-mail: thomas.geiser@insel.ch)}% <-this % stops a space
\thanks{L. Ebner and A. Christe are with the Department of Diagnostic,
Interventional and Pediatric Radiology, Bern University Hospital ``Inselspital'',
3010 Bern, Switzerland (e-mails: lukas.ebner@insel.ch; andreas.christe@insel.ch).}% <-this % stops a space
\thanks{S. Mougiakakou* is with the Department of Diagnostic, Interventional
and Pediatric Radiology, Bern University Hospital ``Inselspital'', 3010
Bern, Switzerland, and the ARTORG Center for Biomedical Engineering Research, University of Bern, 3008 Bern Switzerland (e-mail:
stavroula.mougiakakou@artorg.unibe.ch).}
\thanks{Source code available at: https://github.com/intact-project/LungNet}}


\maketitle

\begin{abstract}

Early and accurate diagnosis of interstitial lung diseases (ILDs) is crucial for making treatment decisions, but can be challenging even for experienced radiologists. The diagnostic procedure is based on the detection and recognition of the different ILD pathologies in thoracic CT scans, yet their manifestation often appears similar. In this study, we propose the use of a deep purely convolutional neural network for the semantic segmentation of ILD patterns, as the basic component of a computer aided diagnosis (CAD) system for ILDs. The proposed CNN, which consists of convolutional layers with dilated filters, takes as input a lung CT image of arbitrary size and outputs the corresponding label map. We trained and tested the network on a dataset of 172 sparsely annotated CT scans, within a cross-validation scheme. The training was performed in an end-to-end and semi-supervised fashion, utilizing both labeled and non-labeled image regions. The experimental results show significant performance improvement with respect to the state of the art.

\end{abstract}


%%%%%%%%%%%%%%%%%%%%%%%%%%%%%%%%%%%%%%%%%%%%%%%%%%%%%%%%%%%%%%%%%%%%%%%%%
%%%%%%%%%%%%%%%%%%%%%%%%%%%%%%%ADDITIONAL%%%%%%%%%%%%%%%%%%%%%%%%%%%%%%%%


% \fi
% \IEEEpeerreviewmaketitle
%%%%%%%%%%%%%%%%%%%%%%%%%%%%%%%%%%%%%%%%%%%%%%%%%%%%%%%%%%%%%%%%%%%%%%%%%
%%%%%%%%%%%%%%%%%%%%%%%%%%%%%%%%%%%%%%%%%%%%%%%%%%%%%%%%%%%%%%%%%%%%%%%%%
%%%%%%%%%%%%%%%%%%%%%%%%%%%%%%%%%%%%%%%%%%%%%%%%%%%%%%%%%%%%%%%%%%%%%%%%%

\section{Introduction}

\IEEEPARstart{I}{nterstitial} lung disease (ILD) is a group of more than 200 chronic lung disorders characterized by inflammation and scarring of the lung tissue that leads to respiratory failure. ILD accounts for 15 percent of all cases seen by pulmonologists and can be caused by autoimmune disease, genetic abnormalities, infections, drugs or long-term exposure to hazardous materials. In many cases the cause remains unknown and the disease is described as idiopathic. The diagnosis of ILD is mostly performed by radiologists and is usually based on the assessment of the different ILD pathologies in high resolution computed tomography (HRCT) thoracic scans. Early diagnosis is crucial for making treatment decisions, while misdiagnosis may lead to life-threatening complications~\cite{society1999diagnosis}. Extensive research has been conducted on the development of computer-aided diagnosis (CAD) systems, which are able to support clinicians and improve their diagnostic performance. The basic characteristics of such a system are the automatic detection and recognition of the pathological lung tissue. Pathological tissue is usually manifested as various textural patterns in the CT scan. This ILD pattern recognition procedure is traditionally performed by a local texture classification scheme that slides across the images and outputs a map of pathologies, which is later used to reach a final diagnosis. For texture recognition, a great variety of handcrafted image features and machine learning classifiers have been utilized.

Recently, deep artificial neural networks (ANN) and, in particular, deep convolutional neural networks (CNNs) have gained a lot of attention after their impressive results in the ImageNet Large Scale Visual Recognition Competition in 2012~\cite{ILSVRC15}. Networks of this kind have existed for decades~\cite{lecun98}, but have only recently managed to achieve adequate performance mainly due to the large volumes of available annotated data, the  massive parallelization capabilities of GPUs and a few design tricks. The potential benefits of deep learning techniques in medical image analysis have also been investigated recently and the first results have been promising~\cite{greenspan2016tmiguest}. In~\cite{AnthimoTMI} , we designed, trained and tested a deep CNN as a fixed-scale, local texture classifier that outperformed traditional methods in ILD pattern classification. However, training large networks on medical images can often be challenging due to the lack of databases that are adequately sized to satisfy the needs of these models. Medical data are scarce and collecting them is a difficult and time consuming process, while their annotation has to be performed by multiple specialists to ensure its validity. To this end, in~\cite{christodoulidis2017multi}, we investigated the potential use of general texture image databases, in a multi-source transfer learning scheme that yielded significant improvements in performance. One remaining limitation in these works is the local nature of the classifier that requires rigorous scanning of the input image with a sliding window, and simultaneous aggregation of the results in the output. This classification scheme can be significantly time-consuming while it may also ignore less local but useful information, if the input size is not appropriately configured.

In this study, we propose the use of a deep fully-convolutional network for the problem of ILD pattern recognition that uses dilated convolutions and is trained in an end-to-end and semi-supervised manner. The proposed CNN takes as input a lung HRCT image of arbitrary size and outputs the corresponding label map, thus avoiding the limitations of a sliding window model. Additionally, the utilization of non-labeled image regions in the learning procedure, permits robust training of larger models and proves to be particularly useful when using databases with sparse annotations.

 %While last, we share the code that was used for the purposes of this paper at %https://github.com/intact-project/.

\section{Related Work}

% intro
In this section, we provide a short review of the recent advances in deep learning for computer vision, followed by a brief overview of previous studies on ILD pattern classification

\subsection{Deep CNNs for Computer Vision}

% DL early years
Although CNNs have existed for decades~\cite{lecun98}, they only became widely popular after their remarkable success in the ImageNet challenge of 2012~\cite{ILSVRC15}. The winning approach of the competition~\cite{krizhevsky2012imagenet}, also known as AlexNet, was a deep CNN with five convolutional and three dense layers, each followed by a rectified linear unit (ReLU), while increased strides were used for the max pooling and convolution operations to gradually down-sample the feature maps. Dropout~\cite{Dropout} and data augmentation were also utilized, in order to prevent overfitting. Since then, the proposed deep CNNs have led to continuous improvements in the results on ImageNet and other datasets, mainly by enhancing their architecture and by increasing their depth and width. The VGG network~\cite{simonyan2014very} reduced the size of the kernels to 3$\times$3, while increasing the number of layers to 19. GoogleNet~\cite{szegedy2015going} used consecutive inception modules, where different convolutional and pooling operations are performed in parallel with their outputs merged. This approach drastically reduced the number of parameters while improving the results. ResNet~\cite{he2016deep} introduced skip connections between layers which permitted the training of networks with hundreds of layers and pushed the limits of deep CNNs even further. The batch normalization (BN) technique~\cite{IoffeBN} also supported these developments, by regularizing and accelerating the training procedure.

Many of the already proposed CNNs have recently been adapted to perform semantic segmentation, rather than just image classification. The term semantic segmentation refers to the task of assigning a class label to every pixel of an image. A simple approach to this task is to use any fixed-scale classification method under a sliding window scheme and then to aggregate the results to build a label map. However, this could be highly inefficient as local image features would have to be recalculated multiple times for adjacent positions of the input window. Luckily, the convolutional layers (with the appropriate padding) in a typical CNN produce feature maps that maintain spatial correspondence with the input image. Therefore, input images of any size can be fed to the network and each pixel can be classified, on the basis of the values of the respective feature map position. This can be achieved by utilizing convolutional layers of size 1 $\times$ 1 that serve as local dense layers and, these networks are therefore often referred to as fully convolutional (FCNs). 

However, the spatial correspondence between the input and output of a CNN, can be disrupted by the use of down-sampling operations such, as strided pooling and convolution. Down-sampling of the feature maps is often used to increase the receptive field of the network with respect to the input, as well as to reduce the amount of computational load. In order to restore the original size of the input, researchers have used encoder-decoder architectures, where the encoder usually adopts a well-known architecture such as VGG~\cite{simonyan2014very} and the decoder reverses the process by mapping the feature representation back to the input data space. To this end, upsampling operations and transposed convolution ~\cite{zeiler2014visualizing} (also known as fractionally strided convolution or ``deconvolution'') have been used for semantic segmentation~\cite{long2015fully}. Alternatively, in~\cite{badrinarayanan2015segnet} and~\cite{noh2015learning}, max unpooling has been used as the inverse operation of each max pooling layer, where the pairs of pooling/unpooling layers are coupled by transferring the max indices from the encoder to the decoder. In~\cite{Unet}, a similar architecture was proposed for biomedical image segmentation, with additional skip connections that concatenate the feature maps of an encoding layer to the feature maps of the same-scale decoding layer.

%% This is the figure for the dilated convolutions
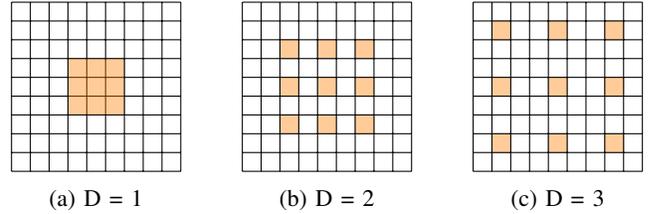
\begin{figure}
    \begin{subfigure}[b]{.15\textwidth}
        \centering
        \begin{tikzpicture}
    
        \draw[step=0.25cm,color=black] (0,0) grid (2+0.25,2+0.25);    
        \draw[fill=orange, opacity=0.4] (0.75,0.75) rectangle ++(0.75,0.75);

        \end{tikzpicture}
        \caption{D = 1}
    \end{subfigure}
    ~
    \begin{subfigure}[b]{.15\textwidth}
        \centering
        \begin{tikzpicture}
        
        \draw[step=0.25cm,color=black] (0,0) grid (2+0.25,2+0.25);    
        
        \draw[fill=orange, opacity=0.4] (0.5,0.5) rectangle ++(0.25,0.25);
        \draw[fill=orange, opacity=0.4] (1,0.5) rectangle ++(0.25,0.25);
        \draw[fill=orange, opacity=0.4] (1.5,0.5) rectangle ++(0.25,0.25);
        
        \draw[fill=orange, opacity=0.4] (0.5,1) rectangle ++(0.25,0.25);
        \draw[fill=orange, opacity=0.4] (1,1) rectangle ++(0.25,0.25);
        \draw[fill=orange, opacity=0.4] (1.5,1) rectangle ++(0.25,0.25);
        
        \draw[fill=orange, opacity=0.4] (0.5,1.5) rectangle ++(0.25,0.25);
        \draw[fill=orange, opacity=0.4] (1,1.5) rectangle ++(0.25,0.25);
        \draw[fill=orange, opacity=0.4] (1.5,1.5) rectangle ++(0.25,0.25);
        
        \end{tikzpicture}
        \caption{D = 2}
    \end{subfigure}
    ~
    \begin{subfigure}[b]{.15\textwidth}
        \centering
        \begin{tikzpicture}
        
        \draw[step=0.25cm,color=black] (0,0) grid (2+0.25,2+0.25);    
        
        \draw[fill=orange, opacity=0.4] (0.25,0.25) rectangle ++(0.25,0.25);
        \draw[fill=orange, opacity=0.4] (1,0.25) rectangle ++(0.25,0.25);
        \draw[fill=orange, opacity=0.4] (1.75,0.25) rectangle ++(0.25,0.25);
        
        \draw[fill=orange, opacity=0.4] (0.25,1) rectangle ++(0.25,0.25);
        \draw[fill=orange, opacity=0.4] (1,1) rectangle ++(0.25,0.25);
        \draw[fill=orange, opacity=0.4] (1.75,1) rectangle ++(0.25,0.25);
        
        \draw[fill=orange, opacity=0.4] (0.25,1.75) rectangle ++(0.25,0.25);
        \draw[fill=orange, opacity=0.4] (1,1.75) rectangle ++(0.25,0.25);
        \draw[fill=orange, opacity=0.4] (1.75,1.75) rectangle ++(0.25,0.25);

        \end{tikzpicture}
        \caption{D = 3}
    \end{subfigure}
    \caption{Dilated convolution kernels. D denotes the dilation rate}
    \label{fig:dilation}
\end{figure}

Recently, some CNNs for semantic segmentation have been proposed that use dilated convolutions to increase the receptive field, instead of downsampling the feature maps. Dilated convolution, also called \`{a}trous, is the convolution with kernels that have been dilated by inserting zero holes (\`{a}trous in French) between the non-zero values of a kernel. This was originally proposed for efficient wavelet decomposition in a scheme also known as ``algorithme \`{a}trous''~\cite{atrous}.  Figure~\ref{fig:dilation} shows examples of kernels with different dilation rates. Dilated convolution can increase the receptive field without increasing the number of parameters, as opposed to normal convolution. Moreover, feature maps are densely computed on the original image resolution without the need for downsampling. In~\cite{YuDilatedSegm}, a CNN module with dilated convolutions was designed to aggregate multiscale contextual information and improve the performance of state-of-the-art semantic segmentation systems. The module has eight convolutional layers with exponentially increasing dilation rates (i.e. 1, 1, 2, 4, 8, 16), resulting in an exponential increase in the receptive field, while the number of parameters is only grown linearly. Similarly, expansion of the receptive field was achieved in~\cite{Enet} by integrating dilated convolutions in a bottleneck module that was designed for efficiency. In~\cite{chen2016deeplab}, the \`{a}trous spatial pyramid pooling (ASPP) scheme is proposed that uses multiple parallel dilated convolutional layers, in order to capture information from multiple scales.

\subsection{ILD Pattern Classification}

Over the last twenty years, numerous approaches have been proposed for the problem of ILD pattern recognition, which is generally regarded as a texture classification problem. Most of the proposed methods involve hand-crafted texture features which are fed to machine learning classifiers, and locally recognize lung tissue within a sliding window framework. In one of the early studies~\cite{uppaluri1999computer}, the adaptive multiple feature method (AMFM) was proposed, which utilizes a combination of gray-level histograms, co-occurrence and run-length matrices, as well as fractal analysis parameters. For the classification, a Bayesian classifier was used. In~\cite{sluimer2003computer}, a filter bank of Gaussian and Laplacian kernels was applied on the input images and the histogram moments of the responses were fed to a linear discriminant classifier. The simple, yet powerful, Local Binary Pattern (LBP) descriptor has also been proposed~\cite{sorensen2010quantitative}, combined with a k-nearest neighbors classifier. In~\cite{antimopoulos2014dct}, a random forest classifier was utilized that was trained on local DCT features. More recently, some proposed methods have adopted unsupervised feature extraction techniques, such as bag of features~\cite{gangeh2010texton, foncubierta2011using} and sparse representation models~\cite{zhao2013classification, vo2011multiscale}. 

Lately, a few methods have been proposed that utilize CNNs for lung pattern classification. Most of these are still designed under a patch-wise scheme where a square patch is fed to the CNN, while the output consists of the probabilities for this patch to belong to each class. Although the strong descriptive capabilities of modern CNNs are commonly attributed to their depth, the first studies utilized rather shallow architectures. A modified RBM that resembles a convolutional layer was used in~\cite{van2014learning}, whereas in~\cite{li2014medical}, a CNN was proposed with one convolutional and three fully-connected layers. More recently, some attempts have also been made to utilize deeper architectures. In our previous work~\cite{AnthimoTMI}, a CNN with five convolutional and three dense layers was designed and trained on ILD data, while in~\cite{christodoulidis2017multi} its performance was improved using knowledge transfer from other domains. In another study~\cite{wang2017multi}, a CNN with three convolutional and one dense layer was fed with rotational invariant Gabor-LBP representations of lung tissue patches. Finally, in~\cite{shin2016deep} and~\cite{gao2017holistic}, the authors utilized well established pretrained CNN architectures such as AlexNet and GoogleNet which were further finetuned for detecting possible ``presence/absence'' of pathologies at a slice level. However, these architectures were designed to classify natural color images with size 224 $\times$ 224, so the authors had to resize the images and artificially generate three channels by applying different Hounsfield unit (HU) windows.
\section{Materials and Methods}
This section presents the proposed fully convolutional neural network for semantic lung tissue segmentation. Prior to this, we describe the materials used for training and testing the network.

\subsection{Materials}
For the purposes of this study, we compiled a dataset of 172 HRCT scans, each corresponding to a unique ILD or healthy subject. The dataset contains 109 cases from the publicly available multimedia database of interstitial lung diseases~\cite{depeursinge2012building} by the Geneva University Hospital (HUG), along with 63 cases from Bern University Hospital - ``Inselspital'' (INSEL), as collected by the authors. The scans were acquired between 2003 and 2015 using different scanners and acquisition protocols. The INSEL scans are volumetric, while the HUG scans have a 10-15mm spacing. The slice thickness is 1-2mm for both datasets.

% INSEL DB acquisition study date interval 2006-2015
% HUG DB acquisition study date interval 2003-2008

\begin{figure}

    \centering
        \begin{subfigure}{3.2in}
        \includegraphics[width=3.2in]{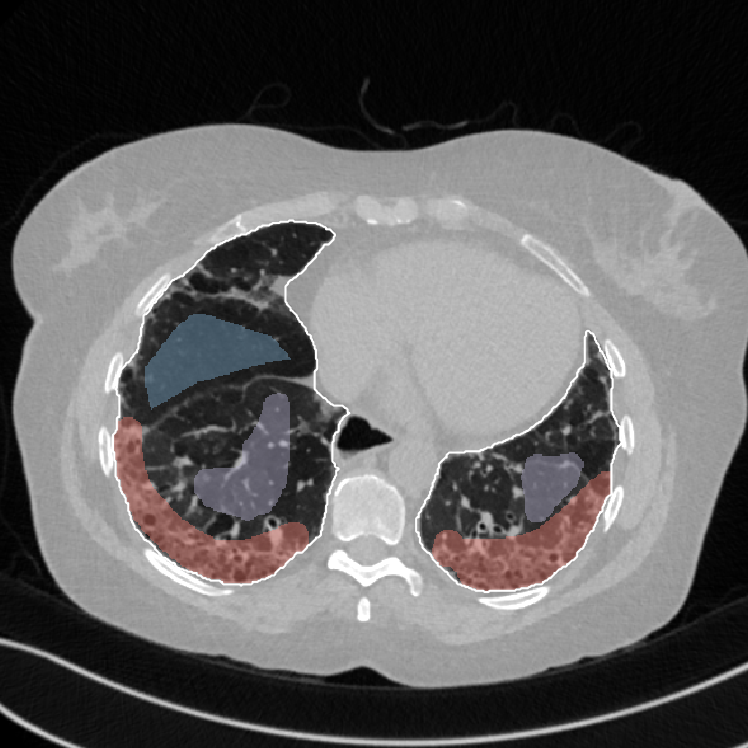}
    \end{subfigure}
    \par\medskip
    \begin{subfigure}{2.5in}
        \centering
        \includegraphics[width=2.5in]{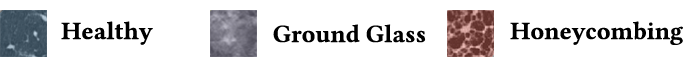}
    \end{subfigure}
    ~
    
    \caption{A typical slice with annotations. The white border line denotes the lung field segmentation, the blue denotes healthy tissue, the purple micronodules and the red the honeycombing pattern.}
    \label{fig:slice}
\end{figure}

Two experienced radiologists from INSEL annotated or re-annotated ILD typical pathological patterns, as well as healthy tissue in both databases\footnote{ITK-snap was used for the annotation process, http://www.itksnap.org}. A lung field segmentation mask was also provided for each case. In total six types of tissue were considered: normal, ground glass opacity, micronodules, consolidation, reticulation and honeycombing. It should be emphasized that these annotations do not cover the entire lung field, but only the most typical manifestations of the listed ILD patterns. This protocol was followed in both databases since it permits the annotation of more scans for the same effort and thus increases data diversity. On the other hand, sparse annotations also introduce challenges. Non-annotated lung areas have to be excluded from both supervised training and evaluation. Another challenging characteristic of the databases is the uneven distribution of the considered classes across the cases. Table~\ref{tab:db-stats} provides statistics for the entire dataset, while figure~\ref{fig:slice} presents a sample CT lung slice along with the given annotations.

\begin{table}
\centering
\caption{Data statistics across the considered classes i.e. Healthy (H), Ground Glass Opacity (GGO), Micronodules (MN), Consolidation (Cons), Reticulation (Ret) and Honeycombing (HC).}
\label{tab:db-stats}
\begin{tabular}{@{}cccccccc@{}}
\toprule
~ & H     & GGO   & MN    & Cons & Ret    & HC    & Totals\\ 
\midrule
\#Pixels$\times10^{5}$  & 92.5  & 27.7  & 35.8  & 7.08 & 28.2   & 20.1  & 211.4\\ 
\#Cases        & 66    & 82    & 15    & 46   & 81     & 47    & 172\\
\bottomrule
\end{tabular}
\end{table}

\subsection{Methods}
In this study, we propose the use of a deep purely convolutional network for the problem of lung tissue semantic segmentation. The network is inspired by~\cite{YuDilatedSegm} and consists of solely convolutional layers that use dilated kernels to increase the receptive field, instead of downsampling the feature maps. This kind of network has been shown to be suitable for similar dense prediction problems that require high resolution precision. The proposed network (Fig.~\ref{fig:architecture}) has 13 convolutional layers and a total receptive field of 287$\times$287. Specifically, each of the first ten layers has 32 kernels of size 3$\times$3 and dilation rates 1, 1, 2, 3, 5, 8, 13, 21, 34 and 55, respectively. We chose not to increase the dilation rates exponentially, as is commonly done, in order to avoid extreme gridding problems that have been reported in several studies~\cite{WangCYLHHC17},~\cite{YuKF17}. Instead, we use the first terms of the Fibonacci sequence as dilation rates; this mitigates the gridding problem by providing a less steep dilation rate increase and thus denser sampling.

The output of the first 10 layers, as well as the input of the network, are concatenated, thus leading to 1+10$\times$32 = 321 feature maps, which are passed through a dropout layer with a rate of 0.5 and fed to the rest of the network. This concatenation is allowed by the lack of pooling layers and the appropriate zero padding for each convolution and brings several benefits. It permits the aggregation of features from all different scales and levels of abstraction, while it also facilitates the flow of gradients thought the network and therefore allows faster training. The last three layers have 1$\times$1 kernels and play the role of locally dense layers that reduce the feature dimensionality for each pixel from 321 to 128, 32 and finally 6, which is the number of classes considered. The output is converted into a probability distribution by the softmax function.

\begin{figure}
\centering
\begin{tikzpicture}
    \begin{scope}
            [node distance=30pt, on grid,>=stealth', rounded corners, inner sep=4pt, text centered, 
    		block/.style={rectangle,draw},
    		arrow/.style={black!90, ->}]
    
    % nodes
    \node[block,fill=green!20,text width=3cm](conv) 
    {Dilated Convolution (k,k), D@x};
    \node[block,fill=red!20,xshift=-20pt,node distance=40pt,below=of conv](bn)
    {InstanceNorm};
    \node[circle,draw,fill=white,xshift=20pt,below=of bn](merge)
    {+};
    \node[block,fill=yellow!20,below=of merge](relu)
    {ReLU};
    
    % arrows
    \draw [arrow]
    (conv) edge (bn)
    (bn) edge (merge)
    (merge) edge (relu)
    [bend left=40]
    (conv) edge (merge);
    
    % background
    \begin{pgfonlayer}{background}
    \node [block, fill=gray!15, inner sep=8pt, dashed,fit=(conv) (bn) (merge) (relu)] {};
    \end{pgfonlayer}
    
    \end{scope} 
\end{tikzpicture}
\caption{The block function of the proposed architecture. (k, k) is the size of the convolution kernel and x is the dilation rate.}
\label{fig:block}
\end{figure}
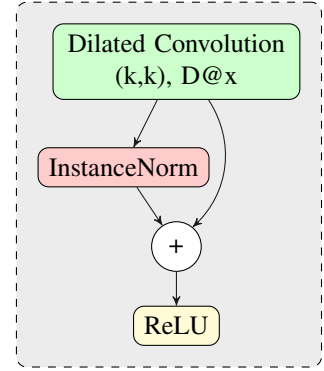

A BN layer follows each convolution and is based on the batch statistics in both training and test time. This is permitted, as the batch size is one, so there is always a full batch during inference. This approach has been proposed before, under the term instance normalization (InstanceNorm)~\cite{TextureNetworks}, and has exhibited good performance in texture synthesis, image stylization and image to image translation~\cite{IsolaZZE16}. InstanceNorm provides invariance to intensity and contrast shifts, which makes the features adaptive for each slice and could mitigate problems caused by different CT scanners and reconstruction kernels. We also found that adding the normalized activations to the non-normalized ones, before passing them through the ReLU function (Fig.~\ref{fig:block}) substantially improves the results. This instance normalization skip connection cancels the mean normalization of activations (when the trainable parameters have not been trained), while it performs a kind of feature contrast enhancement which reduces the importance of variance shift without providing complete invariance to the latter.

The network was trained by minimizing the categorical cross entropy using the Adam optimizer~\cite{Adam} with a learning rate of 0.0001. The dense nature of the considered classification problem combined with the sparse available annotations, resulted in two issues. Firstly, large parts of the dataset were not annotated, and so could not be used for either supervised training or testing. Secondly, the distribution of the considered classes in the dataset was highly imbalanced, a fact that can be challenging for any classification method. We tackled both problems by scaling the considered loss and accuracy with appropriate weighting schemes computed for each set. All pixels corresponding to annotated areas were assigned a weight inversely proportional to the number of samples of its class in the specific set. In this way, all classes contributed equally to the considered metrics. Furthermore, we employed a semi-supervised learning technique to additionally exploit non-labeled areas of the data. We added an extra term to the supervised loss function, which corresponds to the entropy of the network's output on the areas that do not participate in the supervised learning. This entropy minimization technique has been used in different applications such as in ~\cite{entr_min_semi_sup} yielding significant improvements in performance. Similarly in ~\cite{lee2013pseudo} the technique of pseudo-labelling was introduced, where the network classifies non-annotated regions and then uses them as ground truth for fine-tuning. Semi-supervised learning techniques of this kind are based on the cluster assumption i.e. samples from the same class tend to form compact clusters. By minimizing the entropy of the network's output, the decision boundaries are driven away from areas densely populated by learning samples. If the cluster assumption holds and there is no large overlap between the classes this method may increase the network's generalization ability. It acts equivalently to manifold learning and includes self-learning as a special case, as it increases the confidence of the classifier. The influence of the semi-supervised term is controlled by an appropriate weight, which is scaled relatively with the proportion of the unlabeled regions versus the annotated ones. Hence, the loss for a pixel $x$ with output $\mathbf{\hat{y}}$ is:

\begin{equation}
\mathcal{L}(x, \mathbf{\hat{y}}) = 
\left \{ 
\begin{matrix}
-\sum_{i=1}^{C}w_s^{i} y_i\log(\hat{y_i}), & \text{when }\mathbf{y}\text{ is given} \\\\
-\alpha w_u\sum_{i=1}^{C}\hat{y_i}\log(\hat{y_i}), & \text{otherwise}
\end{matrix}
\right.
\end{equation}

where $\mathbf{y}$ is the true label in one-hot encoding, $C$ is the number of classes, $w_s^{i}$ is the supervised weight for class $i$ (which is inversely proportional to the number of samples of the class), $\alpha$ is a scaler and $w_u$ the unsupervised weight.

The training procedure stops when the network does not significantly improve its performance on the validation set for 50 epochs. The performance is assessed in terms of weighted (balanced) accuracy, while an improvement is considered significant if the relative increase in performance is at least 0.5\%. In order to artificially increase the volume of training data and avoid overfitting, we transformed the images using flips and rotations, which are considered label-preserving in this domain. The augmentation was performed online i.e. for each training image in each epoch, one operation out of all eight combinations of flip and rotate is randomly selected and applied.

%% This is the figure with the architecture
\begin{figure}
\centering
\begin{tikzpicture}
    \begin{scope}
            [on grid,>=stealth', rounded corners, text centered,
    		block/.style={rectangle,draw,fill=gray!15,inner sep=5pt},
    		arrow/.style={black!90, ->}]
    		
    % nodes
    \node [block] (input) [fill=white] {Input};
    \node [block] (norm) [below=of input, xshift=-25pt, fill=red!15] {InstanceNorm};
    \node [block] (add) [circle, xshift=25pt, below=of norm, fill=white] {+};
    \node [block] (conv1) [below=of add] {(3x3), D@1};
    \node [block] (conv2) [below=of conv1] {(3x3), D@1};
    \node [block] (conv3) [below=of conv2] {(3x3), D@2};
    \node [block] (conv4) [below=of conv3] {(3x3), D@3};
    \node [block] (conv5) [below=of conv4] {(3x3), D@5};
    \node [block] (conv6) [below=of conv5] {(3x3), D@8};
    \node [block] (conv7) [below=of conv6] {(3x3), D@13};
    \node [block] (conv8) [below=of conv7] {(3x3), D@21};
    \node [block] (conv9) [below=of conv8] {(3x3), D@34};
    \node [block] (conv10) [below=of conv9] {(3x3), D@55};
    
    \node [block] (merge) [below=of conv10, fill=green!15, minimum width=3cm] {Concatenate};
    \node [block] (drop)  [below=of merge, fill=yellow!20] {Dropout};
    
    \node [block] (conv11) [below=of drop] {(1x1), D@1};
    \node [block] (conv12) [below=of conv11] {(1x1), D@1};
    \node [block] (conv13)[below=of conv12] {(1x1), D@1};
    
    \node [block] (output)[below=of conv13,fill=blue!15] {Softmax Output};

    % arrows
    \draw [arrow]
    (input) edge (norm) 
    (norm) edge (add) 
    [bend left=50]
    (input) edge (add)
    [bend right=0]
    (add) edge (conv1)
    (conv1) edge (conv2) 
    (conv2) edge (conv3)
    (conv3) edge (conv4)
    (conv4) edge (conv5)
    (conv5) edge (conv6)
    (conv6) edge (conv7)
    (conv7) edge (conv8)
    (conv8) edge (conv9)
    (conv9) edge (conv10)
    (conv10) edge (merge)
    (merge) edge (drop)
    (drop) edge (conv11)
    (conv11) edge (conv12)
    (conv12) edge (conv13)
    (conv13) edge (output)
    [bend left=40] 
    (input.east) edge (merge.east) 
    (conv1.east) edge (merge.east) 
    (conv2.east) edge (merge.east)
    (conv3.east) edge (merge.east)
    (conv4.east) edge (merge.east)
    (conv5.east) edge (merge.east)
    (conv6.east) edge (merge.east)
    (conv7.east) edge (merge.east)
    (conv8.east) edge (merge.east)
    (conv9.east) edge (merge.east);
    
    \end{scope} 
\end{tikzpicture}
\caption{The architecture of the proposed network. Each gray box corresponds to a block like the one presented in Fig.~\ref{fig:block} }
\label{fig:architecture}
\end{figure}
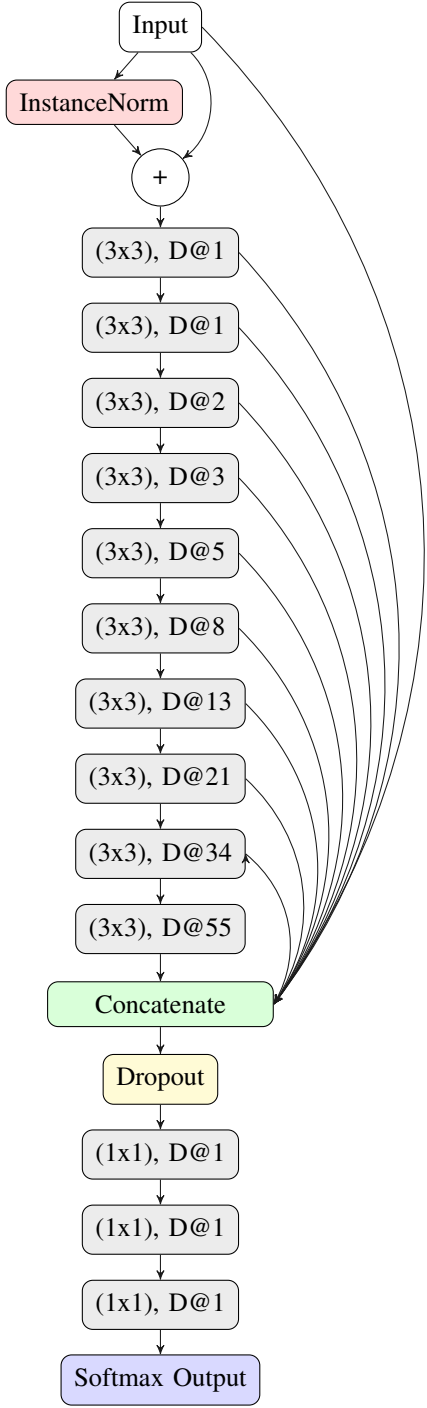

%% This is the figure with the function block

% Experimental setup description (1 page tops) Tables, Figures
\section{Experimental Setup and Results}

In this section, we first present the setup of the experiments conducted, followed by the corresponding results that justify the algorithmic choices of the proposed method and compare it to the state of the art.

\subsection{Experimental Setup}

Given the relatively small size of the dataset with respect to the diversity of the problem, we adopted a 5-fold cross validation (CV) scheme to ensure the validity of the results. The data splitting was performed per scan, so tissue from one case was never present in more than one set. Specifically, the 172 scans of the dataset were divided into five non-overlapping sets, with one of them having 36 and the rest 34 scans. Every time a model was tested on a specific set, the rest of the data were used for training. On average over all folds, the number of slices was 2060 for training and 515 for testing. As principal performance metric, we used the balanced accuracy (Eq.~\ref{eq:1}), averaged over the five folds.

\begin{equation}\label{eq:1}
BACC = \frac{1}{N}\sum_{i=1}^{N}\frac{c_i}{n_i}
\end{equation}

where $N$ is the number of classes, $c_i$ is the number of correctly classified samples of class $i$ and $n_i$ is the total number of samples of class $i$. Since the slices were only sparsely annotated, the accuracy was calculated over the areas of the scans where a ground truth was available. 

In order to avoid extreme class imbalances between the different sets, data splitting was performed using a simple hill climbing technique that maximizes the entropy of the class distribution for the five sets. The methods started from an arbitrary split and then randomly swapped two cases between two sets in an attempt to find a more balanced solution. If the new solution has a higher class distribution entropy (averaged over the 5 sets), we retained it and repeated the procedure until no further improvement was possible.

To minimize the number of computations and memory requirements, we discarded part of the data that lack annotations. Hence, we cropped the left and right lung on each slice and used only the ones with relevant annotations as inputs of the networks. This is permitted by the fully-convolutional nature of the tested networks that do not require a fixed input size. For the cropping, we utilized the available lung mask, while a margin of 32 pixels was added on each side to provide context that could be useful to the networks. 

The proposed method was implemented in Python\footnote{https://github.com/intact-project/LungNet} using the Keras framework\footnote{https://github.com/fchollet/keras} with the Theano~\cite{theano} back-end. All experiments were performed under Linux OS on a machine with CPU Intel Core i7-5960X @ 3.50GHz, GPU NVIDIA GeForce Titan X, and 128 GB of RAM.

\subsection{Results}
Table~\ref{II} presents a comparison between different network configurations. The bold line corresponds to the proposed CNN, while the rest correspond to models that differ from the proposed in only one aspect, as specified in the first column. The rest of the columns provide the number of model parameters, the average inference time per (single-lung) slice, and the average balanced accuracy across the five validation sets. 

\begin{table}[]

\centering
\caption{Comparison of the different network configurations}
\begin{tabular}{@{}llll@{}}

\toprule 
\begin{tabular}[c]{@{}l@{}}Network \\ configuration\end{tabular} 
& \begin{tabular}[c]{@{}l@{}}Number of \\ parameters \\ $\times10^5$\end{tabular} 
& \begin{tabular}[c]{@{}l@{}}Average \\ inference time \\ ms\end{tabular} 
& \begin{tabular}[c]{@{}l@{}}CV balanced\\ accuracy \\ \%\end{tabular} \\ 

\midrule
w/o dilated convolutions                    & 1.30 & 51 &   68.0 \\
w/o concatenation                           & 0.93 & 53 &   72.6 \\
w/o InstanceNorm                            & 1.29 & 38 &   77.9 \\
w/o InstanceNorm skip                       & 1.30 & 57 &   78.6 \\
16 kernels/layer                            & 0.47 & 51 &   79.2 \\
Exponential dilation~\cite{YuDilatedSegm}   & 1.03 & 48 &   79.5 \\
Purely supervised                           & 1.30 & 58 &   80.6 \\
9 dilated layers                            & 1.18 & 53 &   81.3 \\
\textbf{Proposed} & \textbf{1.30} & \textbf{58} &   \textbf{81.8} \\
64 kernels/layer                            & 4.23 & 82 &   82.1 \\

\bottomrule

\end{tabular}
\label{II}
\end{table}

The proposed model achieved top performance with accuracy nearly equal to 82\% and inference time 58ms. The use of 64 kernels per layer instead of 32 did indeed improve the results, yet not significantly enough and with higher inference times, whereas the network with 16 kernels performed notably worse. On reducing the dilated convolutional layers from 10 to 9, we observed a relatively small reduction in the accuracy. However, we chose to keep 10 layers, since the difference in memory and time requirements was also small and because the resulting receptive field was comparable with that of the state of the art networks used for comparison. The use of semi-supervised learning yielded an improvement of nearly 1.5\%, with no additional requirements in computational resources. We also performed an experiment with exponential increase in the dilation rates of the consecutive layers, similarly to~\cite{YuDilatedSegm} i.e. 1, 1, 2, 4, 8, 16, 32 and 64. The resulting model was smaller and faster, since 2 fewer layers were required to achieve similar receptive field, however the accuracy decreased by almost 3\%. In the case where the convolutions were not dilated, the network performed poorly, because of the radical decrease of the receptive field. The accuracy of the proposed model without any normalization was substantially poorer, probably because it could not properly handle the contrast differences among the scans caused by different CT scanners and reconstruction kernels. The use of instance normalization improved the performance by adaptively normalizing the feature contrast for each input. However, this kind of normalization also normalizes the mean intensity that could be a useful feature. By adding the InstanceNorm skip connection (Fig.~\ref{fig:block}), the accuracy improved even further. We speculate this is because the mean normalization is diminished, while the resulting variance normalization is only partially invariant to contrast shifts. Finally, omitting the concatenation of the first 10 layers also resulted in significant impairment of the results, which was expected since only 32 features are considered. 

\begin{figure}
    \centering
    \includegraphics[width=3.5in]{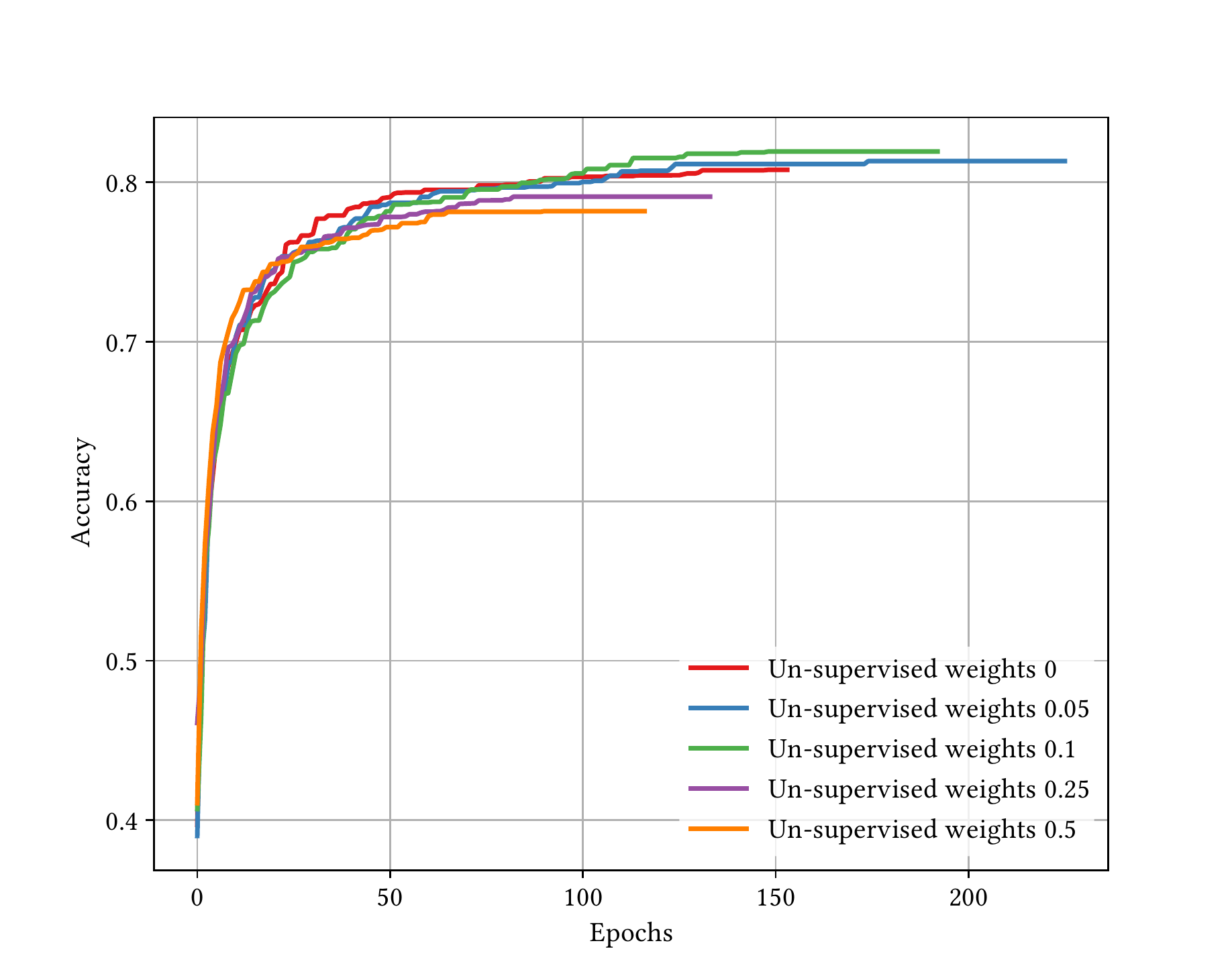}
    \caption{Accuracy curves for different values of $w_u$.}
    \label{fig:curves}
\end{figure}

In Fig.~\ref{fig:curves} the accuracy curves for different values of $w_u$ are presented. These curves are generated by averaging over the five folds the best accuracies achieved this far by each model in each epoch. The curve for the model without the unsupervised learning was also included for comparison. The best performing configuration proved to be the one with $w_u=0.1$, which we utilized for the training of the proposed model.

\begin{figure*}
    \centering
    \begin{subfigure}[b]{1\textwidth}
    	\centering
        \includegraphics[width=1\textwidth]{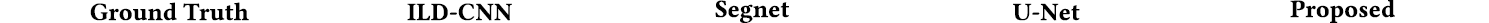}
    \end{subfigure}
    ~
    ~

    \begin{subfigure}[b]{1\textwidth}
    	\centering
        \includegraphics[width=1\textwidth]{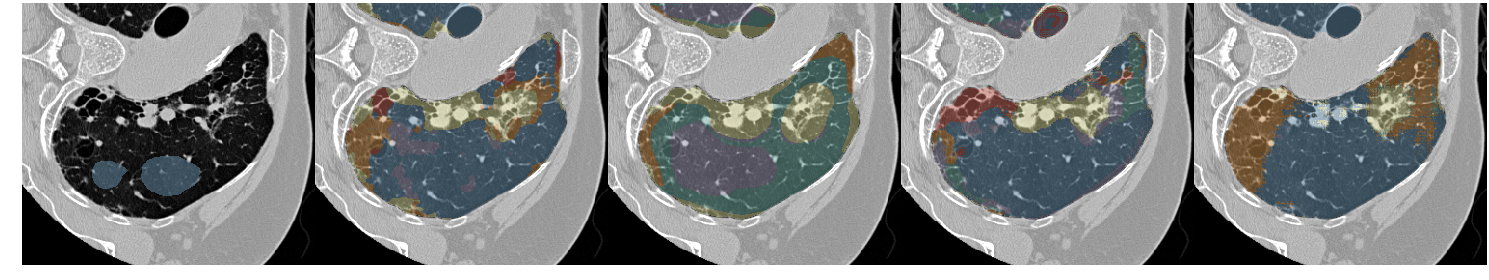}
    \end{subfigure}
    ~
    
    \begin{subfigure}[b]{1\textwidth}
    	\centering
        \includegraphics[width=1\textwidth]{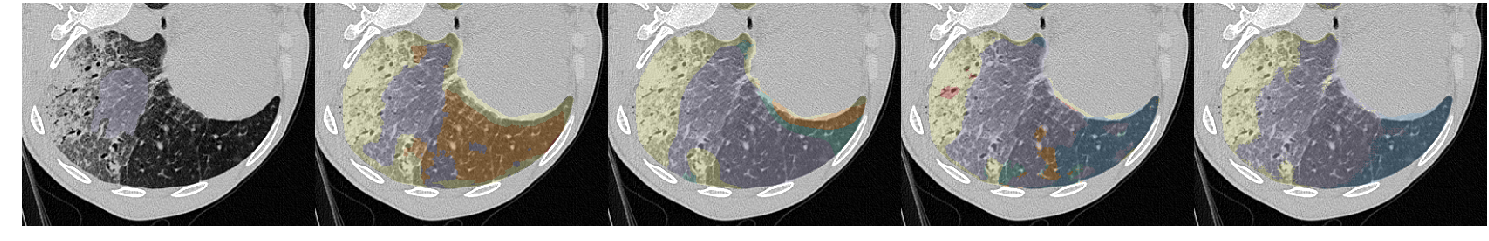}
    \end{subfigure}
    ~
    
    \begin{subfigure}[b]{1\textwidth}
    	\centering
        \includegraphics[width=1\textwidth]{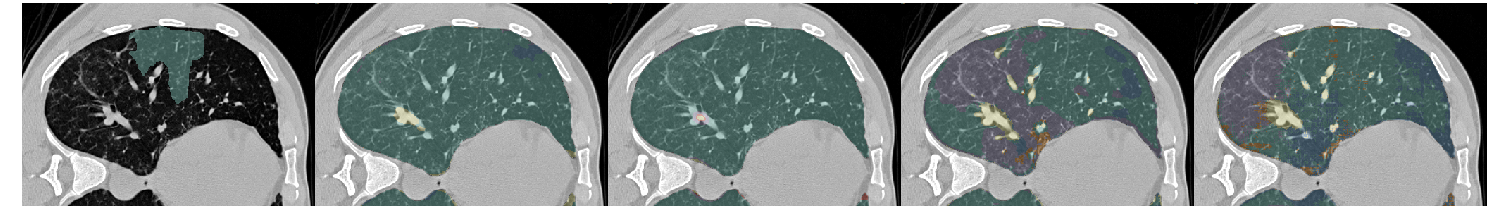}
    \end{subfigure}
    ~
    
    \begin{subfigure}[b]{1\textwidth}
    	\centering
        \includegraphics[width=1\textwidth]{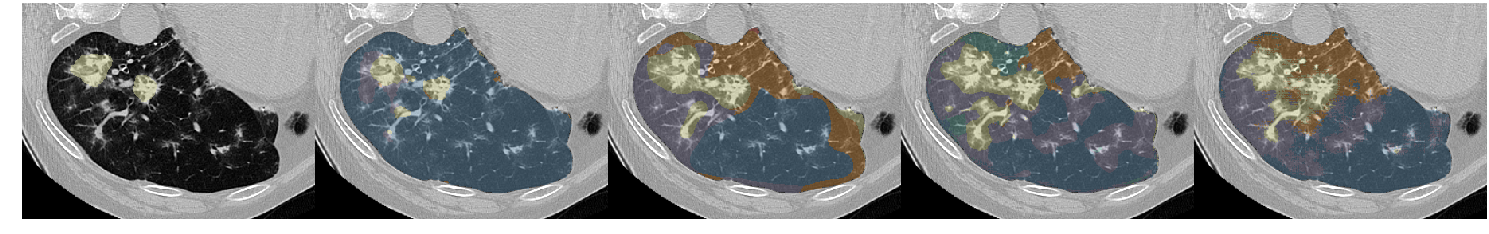}
    \end{subfigure}
    ~
    
    \begin{subfigure}[b]{1\textwidth}
    	\centering
        \includegraphics[width=1\textwidth]{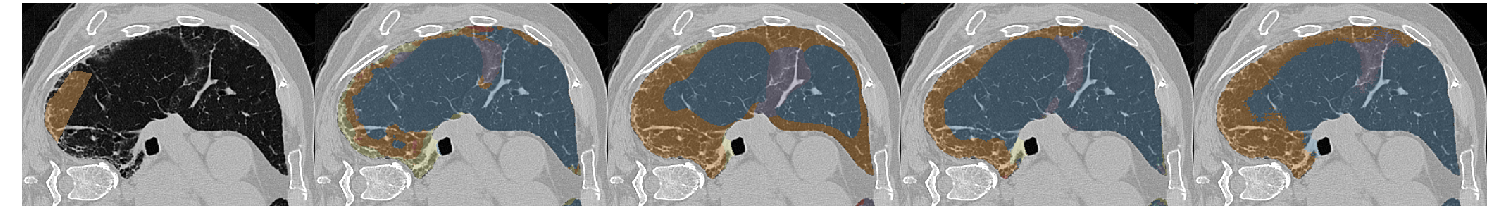}
    \end{subfigure}
    ~

    \begin{subfigure}[b]{1\textwidth}
    	\centering
        \includegraphics[width=1\textwidth]{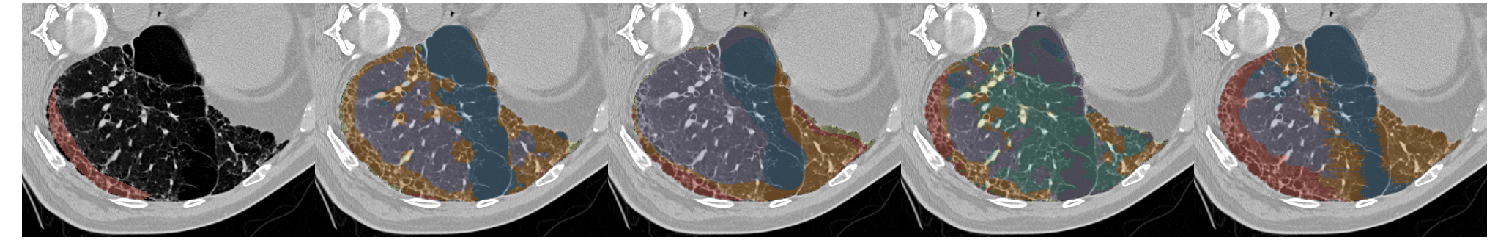}
    \end{subfigure}
    ~
    
    \begin{subfigure}[b]{1\textwidth}
    	\centering
        \includegraphics[width=0.8\textwidth]{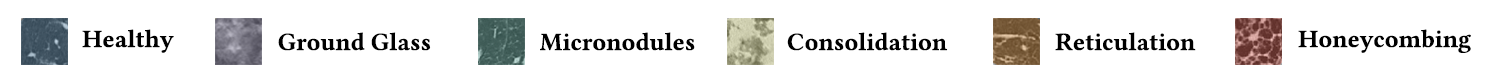}
    \end{subfigure}
    ~
    
    \caption{Output examples for the models of Table~\ref{III}. From left to right: Ground Truth, ILD-CNN, Segnet, U-net, Proposed. Each example has a different pattern annotated. From top to bottom: Healthy (Blue), Ground Glass Opacity (Purple), Micronodules (Green), Consolidation (Yellow), Reticulation (Orange) and Honeycombing (Red).}
    \label{fig:examples}
\end{figure*}

Table~\ref{III} presents a comparison between the proposed network and three previous studies. It has to be noted that all models used the same unsupervised weight ($w_u=0.1$) and whenever batch normalization was performed, this was based on batch statistics (instance normalization) since this yielded the best results. Fig.~\ref{fig:examples} illustrates a few segmentation results for each of the models in Table~\ref{III}.

% \usepackage{booktabs}
\begin{table}[]
\centering
\caption{Comparison with previous studies}
\label{III}
\begin{tabular}{@{}llll@{}}

\toprule 
Network
& \begin{tabular}[c]{@{}l@{}}Number of \\ parameters\\ $\times10^5$\end{tabular} 
& \begin{tabular}[c]{@{}l@{}}Average \\ inference time\\ ms\end{tabular} 
& \begin{tabular}[c]{@{}l@{}}CV balanced\\ accuracy \\ \%\end{tabular} \\ 

\midrule

ILD-CNN~\cite{AnthimoTMI}               & 0.9   & 237 &   72.2 \\
Segnet~\cite{badrinarayanan2015segnet}  & 335   & 111 &   73.6 \\
U-net~\cite{Unet}                       & 310   & 88  &   77.5 \\
\midrule
\textbf{Proposed} & \textbf{1.3}   & \textbf{58}  &   \textbf{81.8} \\ 

\bottomrule
\end{tabular}
\end{table}

\begin{figure}
    \centering
    \includegraphics[width=3in]{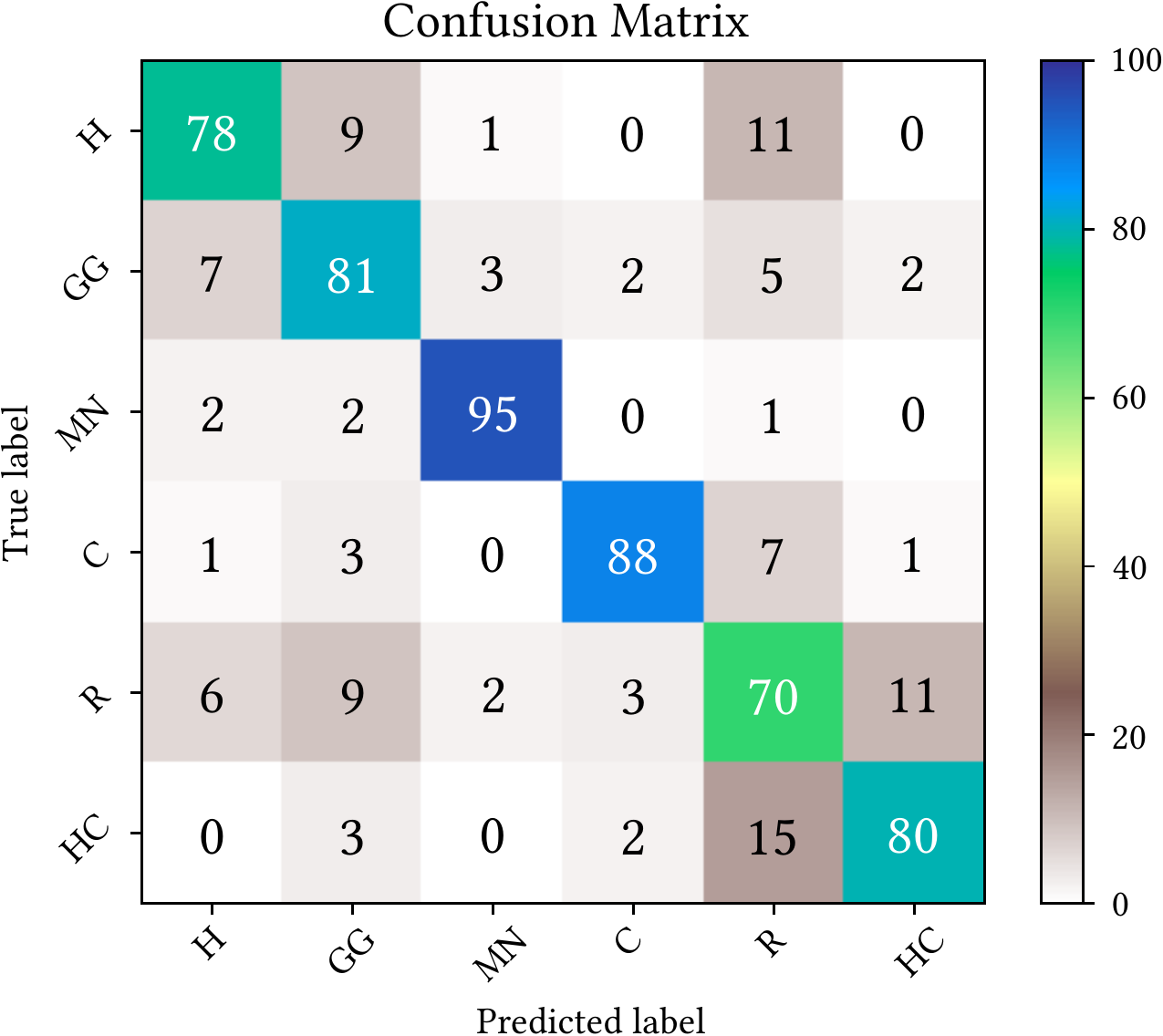}
    \caption{Confusion matrix of the proposed model as calculated over the cross validation scheme. The numbers represent percentages of pixels across all validation images.}
    \label{fig:cm}
\end{figure}

The first line of the table refers to our previous work~\cite{AnthimoTMI}, which has been converted into a fully convolutional network so it can accept arbitrarily sized images for input. Its low accuracy is probably due to the small receptive field (33$\times$33) and the extensive pooling. This architecture was sufficient to describe the local texture of the 32$\times$32 single-class patches in~\cite{AnthimoTMI}, but could not capture higher level structure that is present in the whole-lung dataset of this study. The results of the model in Fig.~\ref{fig:examples} show its noisy output near the lung boundaries or between patterns, where context information could be useful. Segnet~\cite{badrinarayanan2015segnet} and U-net~\cite{Unet} yielded better results, with the latter being slightly faster and substantially more accurate. Both models have a very high number of parameters and large enough receptive fields to capture any relevant information. The superior performance of U-net could be attributed to its skip connections that allow features from the lower scales to directly contribute to its output. Indeed, Fig.~\ref{fig:examples} illustrates the more detailed results of U-net as opposed to the overly smoothed areas produced by Segnet. Finally, the proposed network yielded the best results, while being faster and having far fewer parameters. The output examples in Fig.~\ref{fig:examples} indicate that the proposed model manages to keep a better balance between fine details and smooth border among the different classes. Even thought it is really difficult to visually assess the performance of the system for the different classes, there are a few examples in Fig.~\ref{fig:examples} with wrong classifications on which we can comment. Firstly, parts of the broncho-vascular tree in the third row were recognized as consolidation because of their similar densities, while accentuated terminal bronchial parts, that might be physiological as well, caused the erroneous classification of healthy areas into reticulation, in the first row. Some mistakes however are also attributed in the limited number of annotated classes. For example in row 6, there are emphysematic areas (dark area in the center of the lung) that have been annotated as healthy due to their similar density. Figure~\ref{fig:cm} shows the confusion matrix of the proposed model. As expected, many of the misclassifications occur between reticulation and honeycombing due to their similar textural appearance. Moreover, healthy tissue is often confused with reticulation probably because of the 2D sections of the bronchovascular tree that could resemble reticular patterns.

    % \begin{subfigure}[b]{1\textwidth}
    % 	\centering
    %     \includegraphics[width=0.9\textwidth]{assets/example_12.png}
    % \end{subfigure}
    % ~

% Concluding and small discussion 
\section{Conclusions}
In this study, we proposed and evaluated a deep CNN for the semantic segmentation of pathological lung tissue on HRCT slices. The CNN is designed under a fully convolutional scheme and thus can handle variable input sizes, while it was trained in an end-to-end and semi-supervised fashion. The main characteristic of the proposed network is the use of dilated convolutions along with an instance variance normalization scheme, and multi-scale feature fusion. The training and testing of the network was performed using a cross validation scheme on a dataset of 172 cases, whereas the split of the dataset into folds was performed per case. The proposed network surpassed the highest performance in previous studies, and is much more efficient in terms of memory and computation. Future work includes the modification of the model to consider the 3D nature of lung patterns, and to account for the bronchovascular tree. The former could be achieved by a direct extension of the architecture to 3D, similarly to 3D U-Net~\cite{cciccek20163d} and V-Net~\cite{milletari2016v} or by employing a multi-planar view aggregation scheme, also referred to as 2.5D,~\cite{roth2016improving}. Alternatively, a 3D post processing scheme could be used to refine the 2D segmentation output using conditional random fields or deformation models~\cite{liu2018deep, christ2016automatic, kamnitsas2017efficient}. Finally, the result of a bronchovascular segmentation method could be utilized by the network to reduce false alarms.



\end{document}